\documentclass[10pt,twocolumn]{article}

\usepackage{iccv}
\usepackage{times}
\usepackage{epsfig}
\usepackage{graphicx}
\usepackage{amsmath}
\usepackage{amssymb}
\usepackage{algorithm}
\usepackage{algorithmicx}
\usepackage{algpseudocode}

\graphicspath{{./images/}}

\iccvfinalcopy 

\def\iccvPaperID{3060} 

\newcommand{\bfW}{\mathbf{W}}
\newcommand{\bfZ}{\mathbf{Z}}
\ificcvfinal\fi

\title{Adversarial Examples Detection in Deep Networks with Convolutional Filter Statistics}

\author{Xin Li, Fuxin Li\\
School of Electrical Engineering and Computer Science \\
Oregon State University \\
{\tt\small urumican@gmail.com, lif@eecs.oregonstate.edu}
}

\begin{document}

\maketitle

\begin{abstract}
Deep learning has greatly improved visual recognition in recent years. However, recent research has shown that there exist many adversarial examples that can negatively impact the performance of such an architecture. This paper focuses on detecting those adversarial examples by analyzing whether they come from the same distribution as the normal examples. Instead of directly training a deep neural network to detect adversarials, a much simpler approach was proposed based on statistics on outputs from convolutional layers. A cascade classifier was designed to efficiently detect adversarials. Furthermore, trained from one particular adversarial generating mechanism, the resulting classifier can successfully detect adversarials from a completely different mechanism as well. The resulting classifier is non-subdifferentiable, hence creates a difficulty for adversaries to attack by using the gradient of the classifier. After detecting adversarial examples, we show that many of them can be recovered by simply performing a small average filter on the image. Those findings should lead to more insights about the classification mechanisms in deep convolutional neural networks.

\end{abstract}

\section{Introduction}
\begin{figure}
\includegraphics[width=\columnwidth]{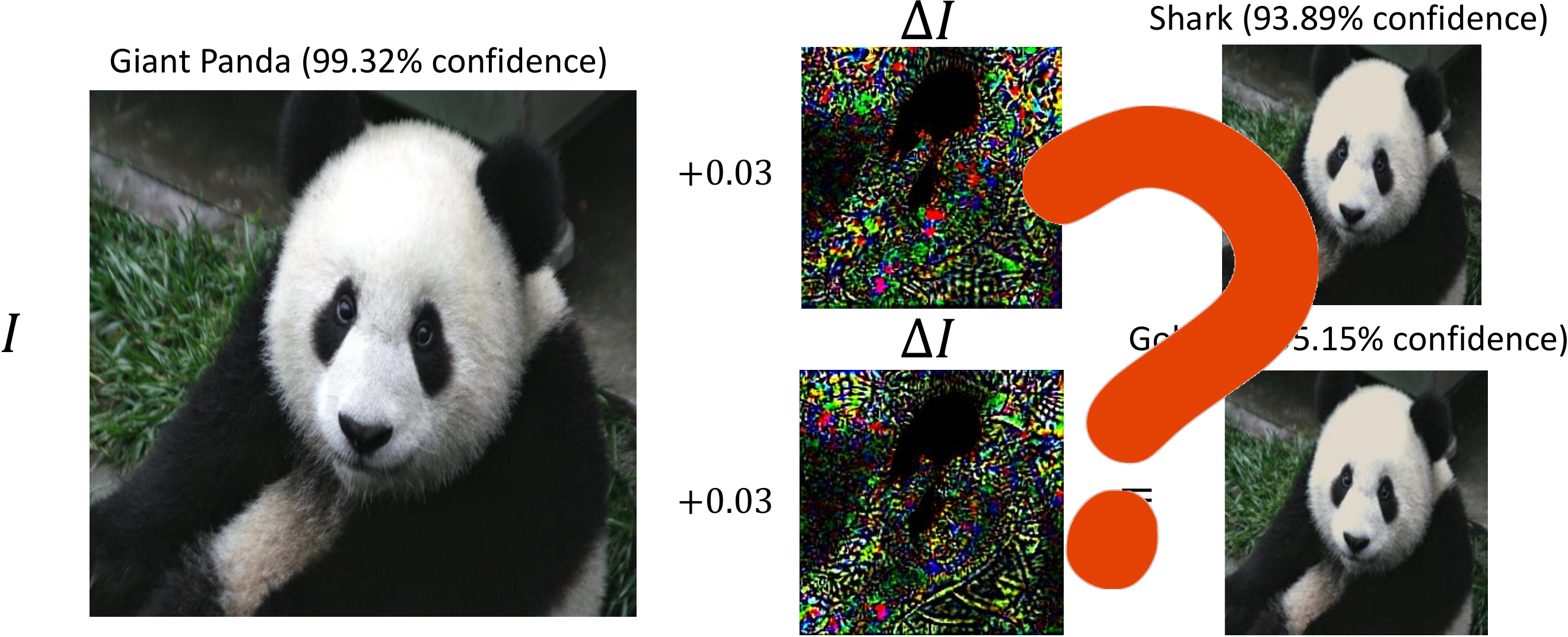}

\caption{An optimization algorithm finds adversarial examples which, with almost negligible perturbations to human eyes, completely distort the prediction result of a deep neural network~\cite{Szegedy2013}. Such algorithms have been found to be universal to different deep networks. This paper studies their properties and seeks a defense.}\label{fig:hard_negatives}
\vskip -0.1in
\end{figure}

Recent advances in deep learning have greatly improved the capability to recognize visual objects~\cite{Krizhevsky2012,Simonyan2014a,He2016residual}. State-of-the-art neural networks perform better than human on difficult, large-scale image classification tasks. However, an interesting discovery has been that those networks, albeit resistant to overfitting, would have completely failed if some of the pixels in the image were perturbed via an adversarial optimization algorithm~\cite{Szegedy2013,Goodfellow2014} . An image indistinguishable from the original for a human observer could lead to significantly different results from a deep network(Fig.~\ref{fig:hard_negatives}). 

Those adversarial examples are dangerous if a deep network is utilized in any crucial real application, be it autonomous driving, robotics, or any automatic identification (face, iris, speech, etc.). If the result of the network can be hacked at the will of a hacker, wrong authentications and other devastating effects would be unavoidable.
Therefore, there are ample reasons to believe that it is important to identify whether an example comes from a normal or an adversarial distribution. A reliable procedure can prevent robots from behaving in undesirable manners  because of the false perceptions it made about the environment. 

The understanding of whether an example belongs to the training distribution has deep roots in statistical machine learning. The \textit{i.i.d.} assumption was commonly used in learning theory, so that the testing examples were assumed to be drawn independently from the same distribution of the training examples. This is because machine learning is only good at performing \textit{interpolation}, where some training examples surround a testing example. \textit{Extrapolation} is known to be difficult, since it is extremely difficult to estimate data labels or statistics if the data is extremely different from any known or learned observations. Many current approaches deal with adversarial examples by adding them back to the training set and re-train. However in their experiments, new adversarials can almost always be found from the re-trained classifier. This is because that the space of extrapolation is significantly larger than the area a machine learning algorithm can interpolate, and the ways to find vulnerabilities of a deep learning system are almost endless.

A more conservative approach is to refrain from making a prediction if the system does not feel comfortable about it. Such an approach seeks to build a wall to fence all testing examples in the extrapolation area out of the predictor, and only predict in the small interpolation area. Work such as~\cite{li2011knows} provides basic theoretical frameworks of classification with an abstain option.

Although these concepts are well-known, the difficulties lie in the high-dimensional spaces that are routinely used in machine learning and especially deep learning. Is it even possible to define interpolation vs. extrapolation in a $4,000$-dimensional or $40,000$-dimensional space? It looks like almost everything is extrapolation since the data is inherently sparse in such a high-dimensional space~\cite{Indyk1998,Hastie2001}, a phenomenon well-known as the curse of dimensionality. The enforcement of the \textit{i.i.d.} assumption seems impossible in such a high-dimensional space, because the inverse problem of estimating the joint distribution requires an exponential number of examples to be solved efficiently. Some recent work on generative adversarial networks proposes using a deep network to train this discriminative classifier~\cite{goodfellow2014generative,radford2015dcgan}, 
where a generative approach is required to generate those samples, but it is largely confined to unsupervised settings and may not be applicable for every domain convolutional networks (CNNs) have been applied to.

In this work we propose a discriminative approach to identify adversarial examples, which trains on simple features and can approach good accuracy with limited training examples. 
The main difference between our approach and previous outlier detection/adversarial detection algorithms (e.g.~\cite{Bendale2016}) is that their approaches usually treat deep learning as a black box and only works at the final output layer, while we believe that the learned filters in the intermediate layers efficiently reduce the dimensionality and are useful for detecting adversarial examples. 
We make a number of empirical visualizations that show how the adversarial examples change the prediction of a deep network. 
From those intuitions, we extract simple statistics from convolutional filter outputs of various layers in the CNN. A cascade classifier is proposed that utilizes features from many layers to discriminate between normal and adversarial examples. 

Experiments show that our features from convolutional filter output statistics can separate between normal and adversarial examples very well. Trained with one particular adversarial generation method, it is robust enough to generalize to adversarials produced from another  generation approach~\cite{Nguyen2015} without any special adaptation or additional training. Those confidence estimates may improve the safety of applying these deep networks, and hopefully provide insights for further research on self-aware learning. As a simple extension, the results from visualizations of the features prompted us to perform an average filter on corrupted  images, and found out that many correct predictions can be recovered from this simple filtering.

\section{Deep Convolutional Neural Networks}

A deep convolutional neural network consists of many convolutional layers which are connected to spatially/temporally adjacent nodes in the next layer:
\begin{equation}
\bfZ_{m+1} = \left[T(\bfW_1 * \bfZ_m), T(\bfW_2 * \bfZ_m), \ldots, T(\bfW_k * \bfZ_m)\right]
\end{equation}
where $\bfZ_m$ is the input features at layer $m$, $\bfW_1, \ldots \bfW_K$ are filters that could be much smaller than the size of $\bfZ_m$ (e.g. $3\times 3, 5\times 5$, $7\times 7$), $*$ is the convolution operator, and $T$ is a nonlinear transformation function such as the rectified linear unit (ReLU) $T(x) = \max(0,x)$.
 Other commonly used layers in a CNN include max-pooling layers, or other normalization layers~\cite{Krizhevsky2012} such as batch normalization layers~\cite{ioffe2015batch}. 
Most deep networks adopt similar principles while adding more structural complexity in the system such as more layers and smaller filters in each layer~\cite{Simonyan2014a}, multi-layered network within each layer~\cite{GoogleLeNet}, residual network~\cite{He2016residual}, etc. 
A convolutional neural network makes sense in structured data because it naturally exploits the locality structure in data. In an image, pixels that are located close to each other are naturally more correlated than pixels that are far away~\cite{li2017filter}. The same holds for temporal data (video, speech) where objects (frames, utterances) that are temporally close can be assumed to be more correlated. 

\section{Understanding the Trained Deep Classifier Under Adversarial Optimization}
\label{sec:understand_trained}
\subsection{Adversarial Optimization}
The famous result that deep networks can be broken easily~\cite{Szegedy2013} is an important motivation of this work.
The idea is to start from an existing example (image) and optimize to obtain an example that will be classified to another category while being close to the original example. Namely, the following optimization problem is solved:
\begin{eqnarray}
\min_r & c\|r\|_1 + L(f_\theta(\mathbf{x}_0 + r, y)) \nonumber \\
 s.t. & \mathbf{x}_0 + r \in [0,1]^d
 \label{eqn:adversarial_optimization}
\end{eqnarray}
where $\mathbf{x}_0$ is a known example and $y$ is an arbitrary category label, $d$ is the input dimensionality. $c$ is a parameter that can be tuned for trading off between proximity to the original example $x_0$ and the classification loss on the other category $y$. It has been shown, to the astound of many, that one can choose an $r$ with very small norm while completely change the output of the algorithm (e.g. Fig.~\ref{fig:hard_negatives}), this can even be done universally for almost all networks, datasets and categories~\cite{Szegedy2013,Goodfellow2014}. Besides, adversarials trained from one network may even fool a related one trained from the same dataset~\cite{Luo2016}. This has led many people to question whether deep networks are really learning the ``proper" rules for classifying those images.

\subsection{Adversarial Behavior}

In order to gain a deeper understanding of the behavior of a deep network and illustrate the difference between adversarial and normal example distributions, we utilize spectral analysis. As a starting point, we perform principal component analysis (PCA)~\cite{PCA} at the 14-th layer of a VGG network trained on the ImageNet dataset (the first fully-connected layer). The rationale behind using PCA is that each deep learning layer is a nonlinear activation function on a linear transformation, hence a large part of the learning process lies within the linear transformation, for which PCA is a standard tool to analyze.

\begin{figure*}[htb]
\begin{tabular}{cccc}
\includegraphics[width=115pt]{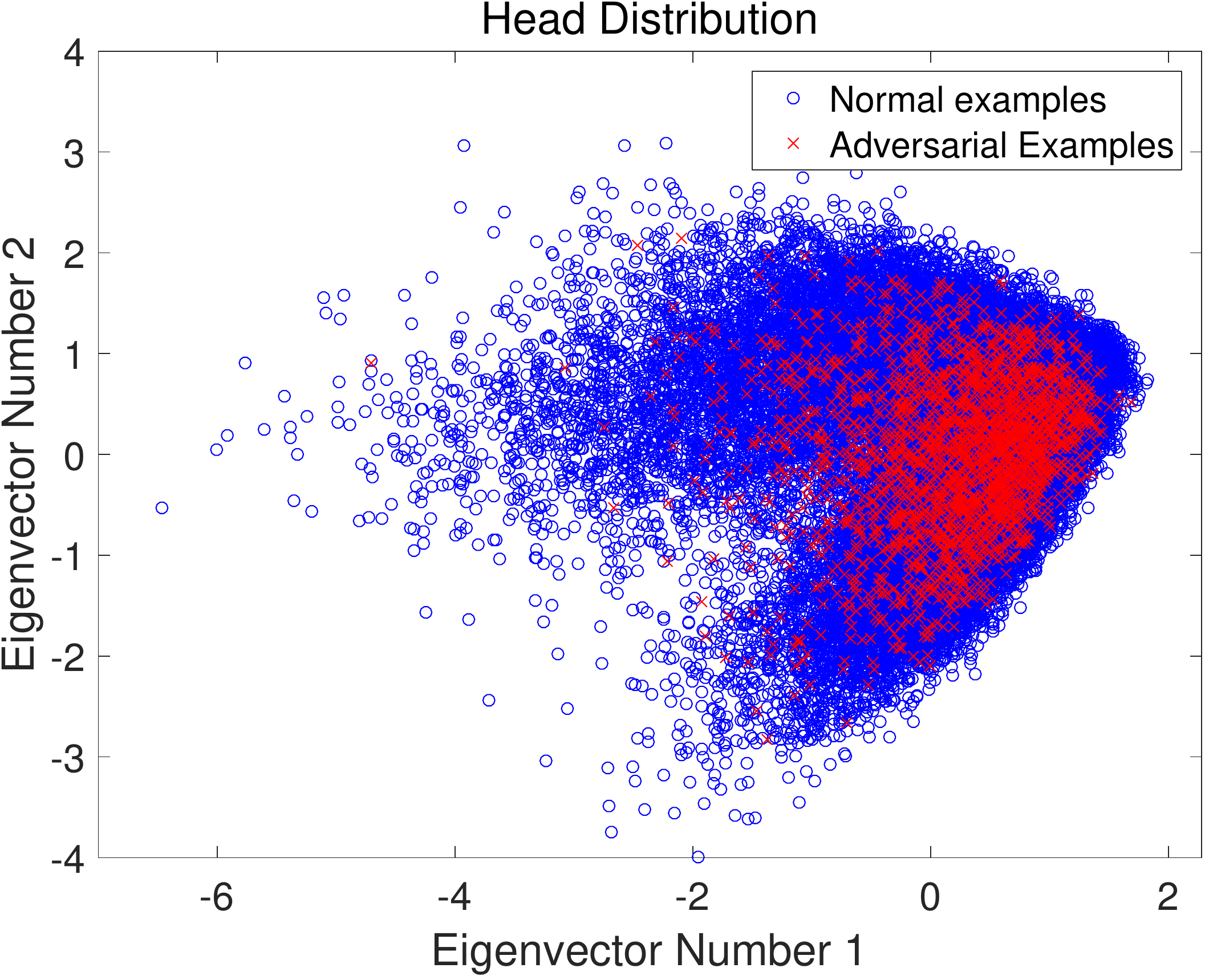} & \includegraphics[width=115pt]{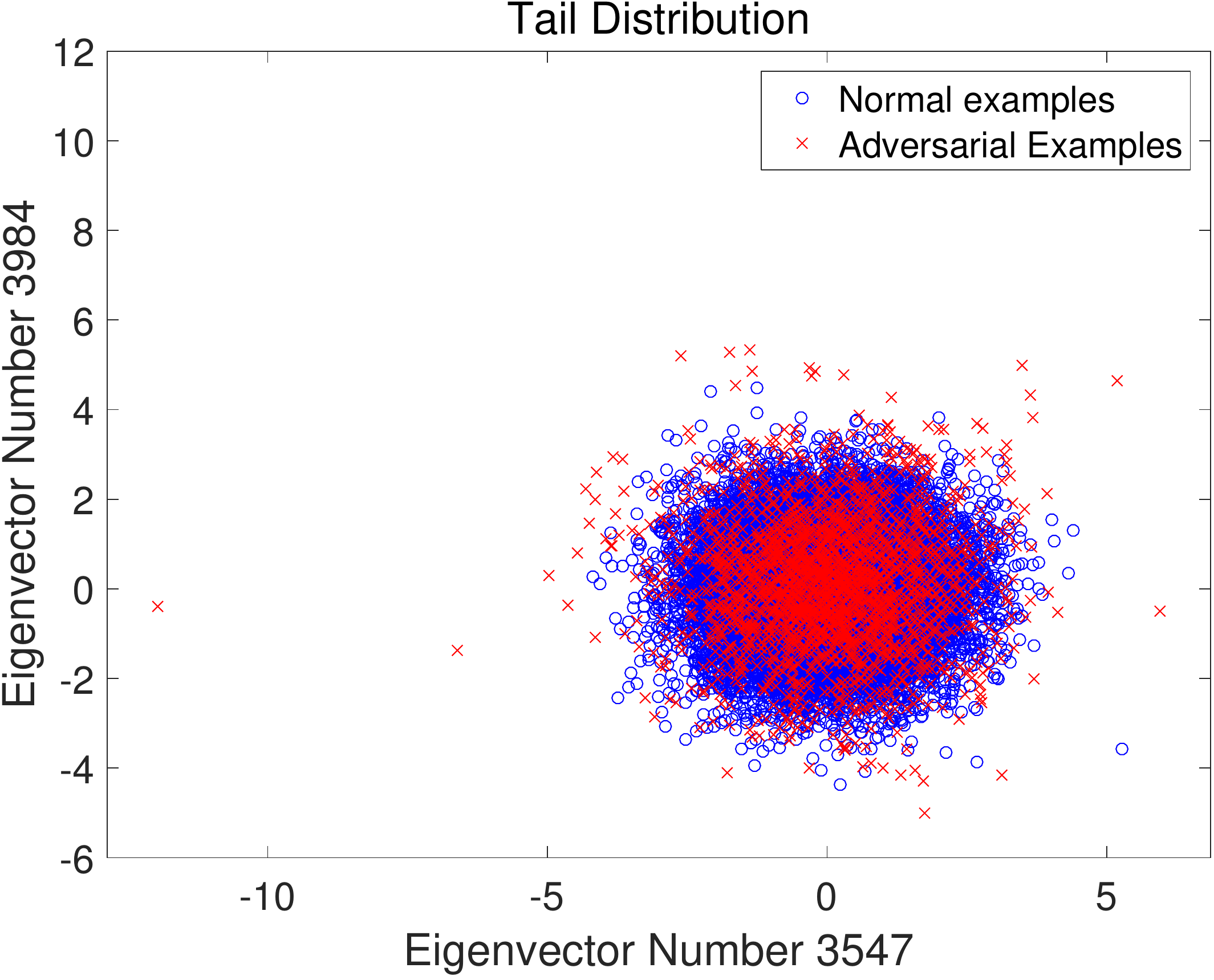} & \includegraphics[width=115pt]{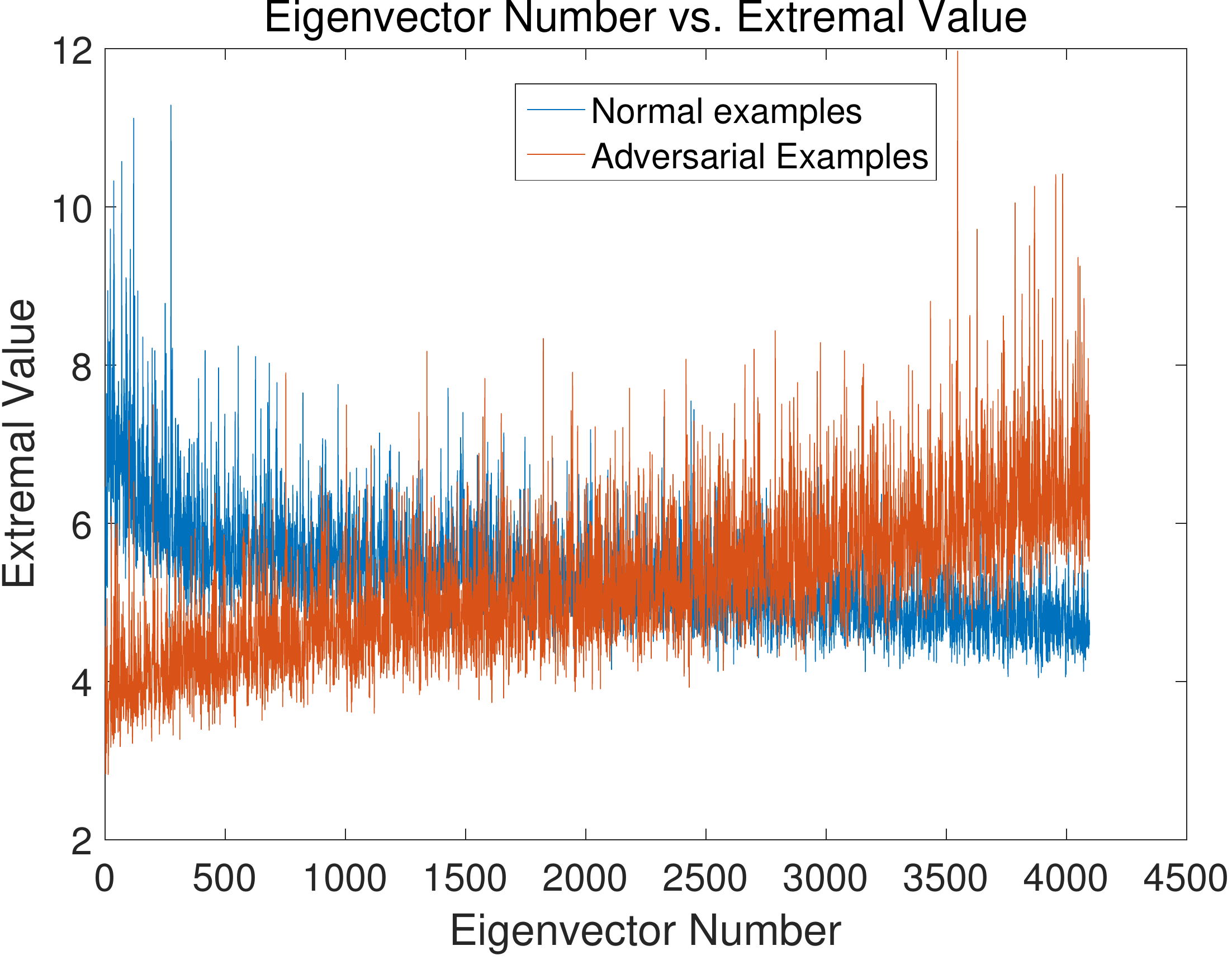} & \includegraphics[width=115pt]{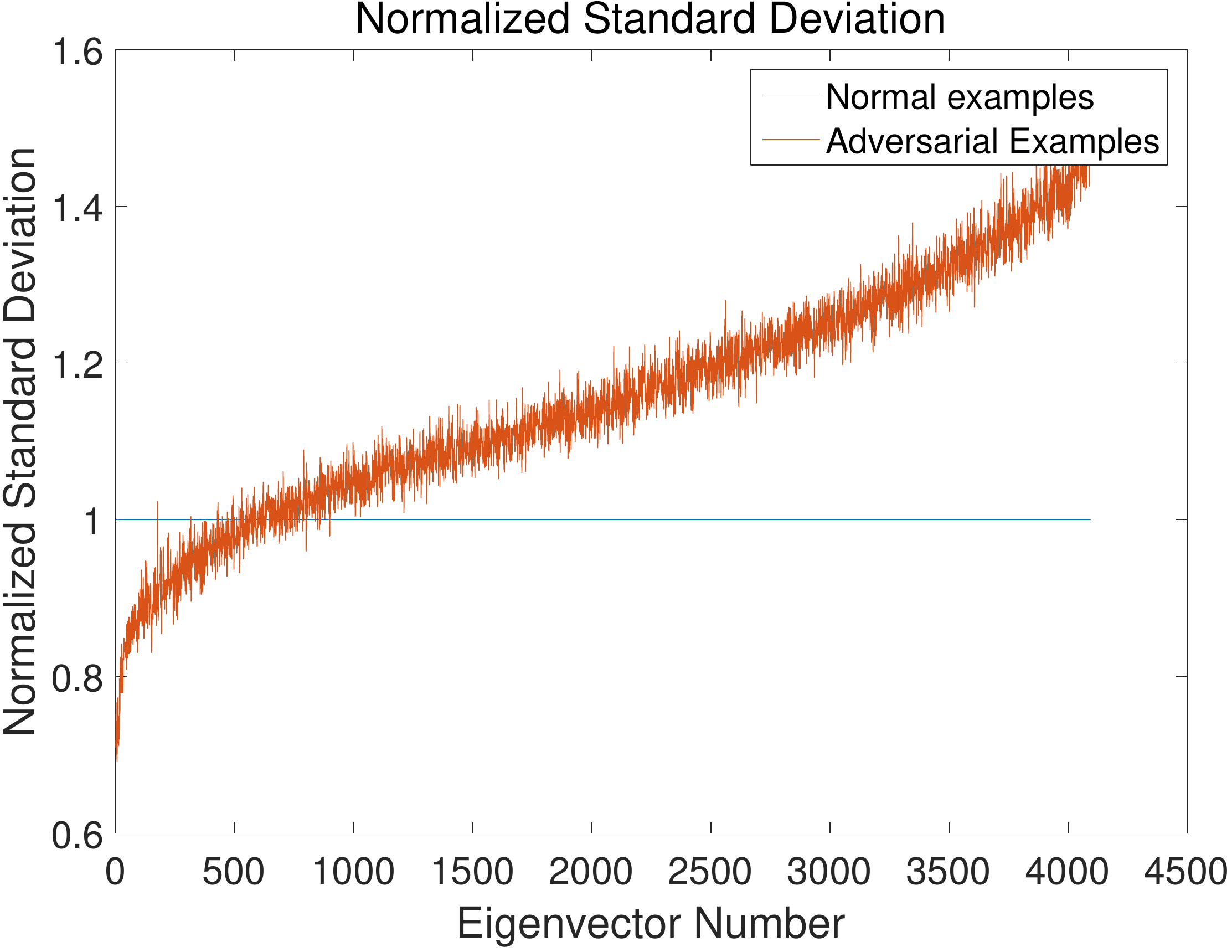} \\
(a) & (b) & (c) & (d)
\end{tabular}
\caption{Blue indicates normal examples and red/orange indicate adversarial examples. Projections are normalized by dividing the standard deviation of all normal examples projected on the corresponding dimension. (a) The projection of the data at layer $14$ onto the 2 most prominent directions; Adversarial example cannot be identified from normal ones. (b) Projection of the same data to the $3,547$-th and $3,844$-th PCA projections, some adversarial examples are having significantly higher deviation to the mean; (c) The absolute normalized extremal value in the projection to each eigenvector; (d) The average normalized standard deviation of normal and adversarial examples on each projection. Standard deviations of normal examples stand at 1 because of the normalization.}
\label{fig:adversarial}
\vskip -0.2in
\end{figure*}

A linear PCA is performed on the entire collection of $50,000$ images from the ImageNet validation set, as well as $4,000$ adversarials collected using the approach in (\ref{eqn:adversarial_optimization}), starting from random images in the collection. The result shows very interesting findings (Fig.~\ref{fig:adversarial}) and sheds more light on the internal mechanics of those adversarial examples. In Fig.~\ref{fig:adversarial}(a), we show the PCA projection onto the first two eigenvectors. This cannot separate normal and adversarial examples, as one could possibly imagine. The adversarial examples seem to exactly belong to the same distribution as normal ones. However, it does seem that the adversarial examples reside mostly in the center while the normal examples occupy a bigger chunk of space.

Interestingly, as we move to the tail of the PCA projection space, the picture starts to change significantly. In Fig.~\ref{fig:adversarial}(b), we can see that there are a significant amount of adversarial examples that has extremely large values w.r.t. to the normal examples in the tail of the distribution. We chose to print the projection on the $3,547$-th and $3,844$-th eigenvector, but similar distributions can be found all over the tail. As one can see, at such a far end on the tail, the projections of normal examples are very similar to random samples under a Gaussian distribution. An explanation for that could be that under these ``uninformative" directions, most of the weighted features are nearly independent w.r.t. each other, hence the distribution of their sum is similar to Gaussian, according to the central limit theorem\footnote{Note this is without a ReLU transformation. ReLU would destroy the negative part of the data distribution so that it no longer looks like a Gaussian. However, some tail effects can be observed even in the distribution after ReLU.}. However, although normal examples behave similarly to a Gaussian, some adversarial examples are having projections with a deviation as large as $5$ or $10$ times the standard deviation, which are extremely unlikely to occur under a Gaussian distribution.

Fig.~\ref{fig:adversarial}(c) and Fig.~\ref{fig:adversarial}(d) show that there are two distinct phenomena:
\begin{itemize}
\vspace{-0.03in}
\item The extremal values and standard deviations on the projections onto the first $500 - 700$ eigenvectors are decidedly lower in adversarial examples than in normal ones.
\vspace{-0.07in}
\item The extremal values and standard deviations on the projections onto the last $1,000 - 1,500$ eigenvectors are decidedly higher in the adversarial examples than the normal ones.
\vspace{-0.03in}
\end{itemize}

\begin{figure}[htb]
\begin{tabular}{cc}
\includegraphics[width=110pt]{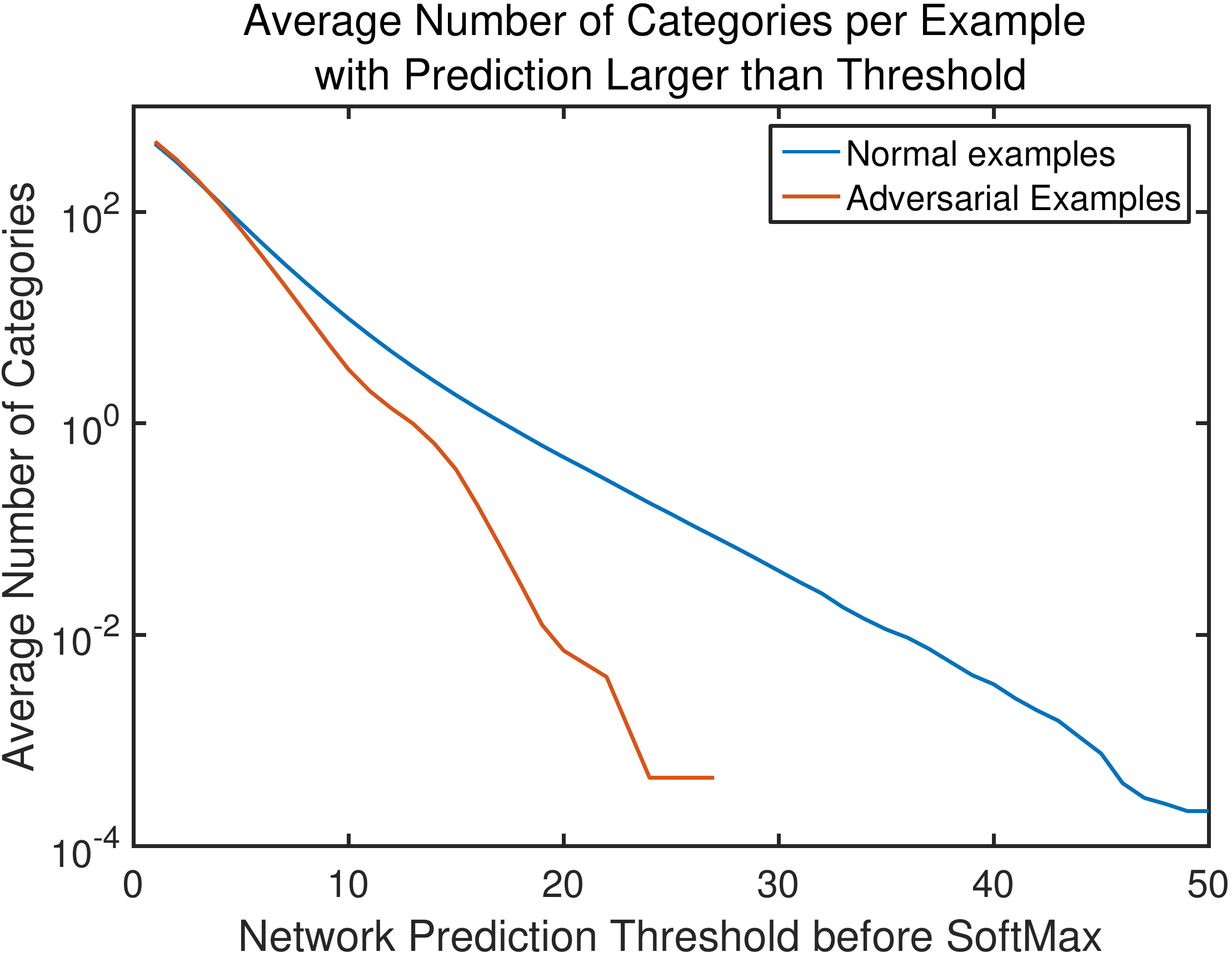} & \includegraphics[width=110pt]{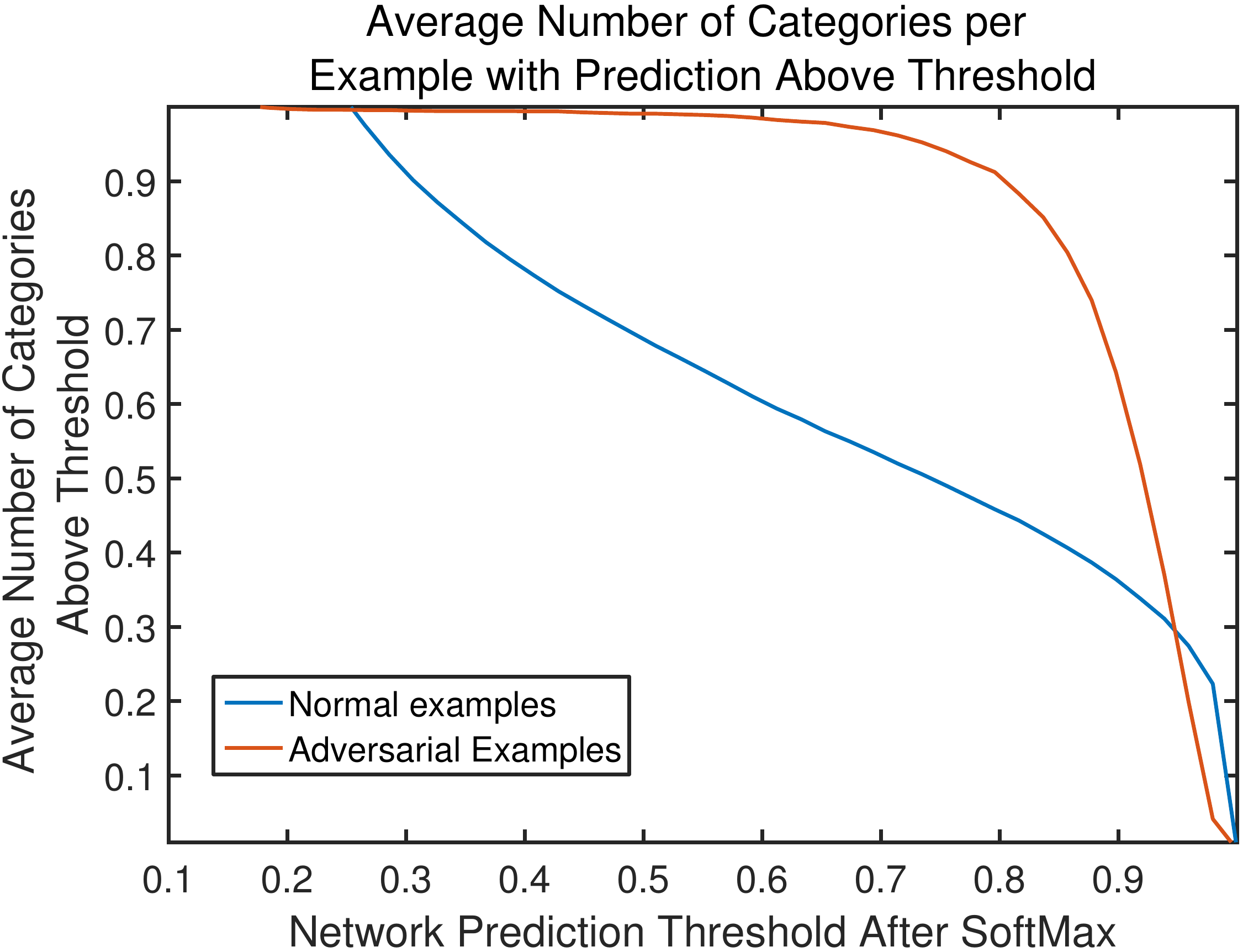}\\
(a) & (b) 
\end{tabular}
\caption{Average number of categories per example with predictions higher than a threshold. (a) Before softmax; (b) After softmax. As one can see, in normal examples, there are on average about $1$ category with a prediction score of more than $20$ (before softmax), while with adversarial ones, only $1\%$ examples have a category with a prediction score more than $20$. However, since prediction values on all categories have dropped, after softmax adversarial examples obtain much higher likelihood on one category.} \label{fig:pred_adversarial}
\end{figure}

It is interesting to reflect about the causes and consequences of those properties. One deciding property is that there is a strong regularization effect in adversarial examples on almost all the informative directions. Hence, the predictions in adversarial examples are \textit{lower} than those in normal examples, rather than the confidence values may have indicated (Fig.~\ref{fig:hard_negatives}). In Fig.~\ref{fig:pred_adversarial}, we show the number of categories with a prediction higher than a threshold, before the final softmax transformation 
\begin{equation}
p_i(x) = \frac{\exp(f_i(x))}{\sum_i \exp(f_i(x))}
\end{equation} that converts raw predictions $f_i(x)$ into probabilities. The result shows that normal examples have on average one category with a raw prediction value more than $20$, however adversarial examples have only $0.01$ category with raw predictions more than $20$. The reason that those adversarial examples appear more \textit{confident} after softmax is because that the predictions on all the other categories are regularized \textit{even more}. Hence the normalization component of softmax has decided that the single prediction, although much less strong, should be assigned a probability of more than $90\%$. We note that this issue was also pointed out by \cite{Bendale2016} in a different manner and they proposed a solution in the OpenMax classifier, which we compare against in the experiments.

But besides that, it seems that such extremal and standard deviation statistics are evident features that could help discriminating normal and adversarial examples. Unfortunately, they only occur as a statistic from a large sample, as any single point in Fig.~\ref{fig:adversarial}(a) looks similar to a single point in the normal distribution. We have tried to utilize the tail distributions (Fig.~\ref{fig:adversarial}(b)) to create a classifier which \textbf{easily achieved more than $99\%$ accuracy separating adversarials from normals}, however we subsequently found out that since the tail almost do not contribute to the classification, knowing this defense, the adversarial example can deliberately remove their footprints on the tail distributions.

This leads us to think about an approach that would turn a single image into a distribution, so that we can use statistics as detectors for adversarial examples. An image is a distribution of pixels. Especially, the output of each filter from each convolutional layer is an image which could be treated as a distribution where the samples are the pixels. Therefore, in the following section we aim to build a classifier based on collecting statistics from such distributions.

\section{Identifying Adversarial Examples}

\subsection{Feature Collection}
\label{sec:features}

Suppose the output at a convolutional layer $m$ is an $W \times H \times K$ tensor, where $W$ and $H$ represent the width and height of the image at that stage (smaller than original after max-pooling), and $K$ represents the number of convolutional filters.  Such a tensor can be considered as a $K$-channel image where each pixel has a $K$-dimensional feature. We consider the feature on every pixel to be a random vector drawn from the distribution $D_m$ of convolutional pixel outputs, a $K$-dimensional distribution. 

The list of statistics we collect is:
\begin{itemize}
\vspace{-0.03in}
\item Normalized PCA coefficients
\vspace{-0.07in}
\item Minimal and Maximal values
\vspace{-0.07in}
\item 25-th, 50-th and 75-th percentile values
\vspace{-0.03in}
\end{itemize}
on each of the $K$-dimensional features. Normalized PCA coefficients are collected via Algorithm~\ref{alg:pca}. Extremal and percentile statistics are straightforward to understand.

The features we collect are non-subdifferentiable, hence essentially preventing adversaries to use gradient-based attacks to counter the classifier. Although we are interested in a generative adversarial network-type adversary which would learn to avoid our detector, such adversaries would have to resort to derivative-free optimization methods, which currently do not scale to the size of a realistic image. The best derivative-free approach we have tried scales up to several hundreds of variables. The genetic algorithm in~\cite{Nguyen2015} scales better, but as we will soon show, their low-level feature statistics are so different from natural images, making them very easy to be detected, even without training on any data from their adversarial generation algorithm.


\begin{algorithm}
\caption{PCA Statistics Extraction} \label{alg:pca}
\begin{algorithmic}[1]
\State \textbf{INPUT:} Image $I$, layer $m$.
\State For all normal images in a training set, compute their CNN filter output of layer $m$ to form an example matrix $\mathbf{Z_m}$.
\State Compute the mean $\mathbf{e}$ and PCA projection matrix $\mathbf{W}$ of $\mathbf{Z_m}$.
\State Compute the standard deviation $\mathbf{s}$ on each dimension in the PCA projection $\mathbf{W}^\top (\mathbf{Z_m} - \mathbf{e1}^\top)$.
\State For each image $I$, project its CNN filter output of layer $m$ $\mathbf{Z_{mI}}$ using PCA: $\mathbf{z_{mI}} = \mathbf{W}^\top (\mathbf{Z_{mI}} - \mathbf{e1}^\top$), and normalize them by dividing the standard deviation $s$ on each respective dimension.
\State Collect the statistic for each image as $\mathbf{x}_I = \frac{1}{n} \|\mathbf{z_{mI}}\|_1$,  where $L_1$ norm is the vector $L_1$ norm.  The resulting statistic is $K$-dimensional.
\end{algorithmic}
\end{algorithm}
%


\subsection{Classifier Cascade}

\begin{figure}[H]
\begin{center}
\includegraphics[width=250pt]{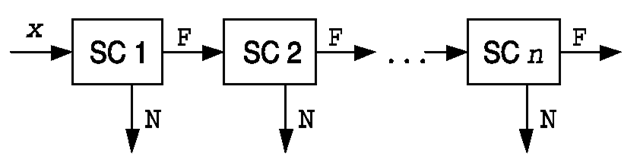}
\vskip -0.3in
\end{center}
\caption{A cascade classifier is defined on each of the convolutional layers in a convolutional network (SC $i$ represents the $i$-th convolutional layer)} \label{fig:cascade_classifier}
\vskip -0.1in
\end{figure}

\cite{viola2004robust} proposed a famous strategy for face detection by using a cascaded boosting classifier composed by a sequence of base classifiers. A cascade classifier is ideal when it is easy to identify many of the examples from a category but some important cases can be difficult. In Fig.~\ref{fig:cascade_classifier}, $SC\_N$ represents the classifier at each stage. $X$ is the input of the cascade classifier. The negatives in a cascade classifier from each stage will be outputted directly, while the positives will go to the next stage.

In our case, the normal category is much easier to detect than the adversarial category (see e.g. Fig.~\ref{fig:overall_roc}). In our initial experiments with VGGNet, we found that more than $80\%$ of normal examples can be determined from the first convolutional layer with $100\%$ precision. Therefore, we constructed a cascade classifier based on convolutional layers: the first stage works with features collected from the outputs of the first convolutional layer, the second with the second layer, etc.~(Fig.~\ref{fig:cascade_classifier}). The base classifiers will not solely consider statistics from their own stage, instead, after one stage of training, the remaining positive examples will be concatenated to the corresponding features on the next stage. 

The operations that are represented by Fig.~\ref{fig:cascade_classifier} can also be summarized as Algorithm~\ref{alg:cascade_train_phase}.
\begin{algorithm}[H]
\caption{Training Process of a cascade of Classifier} \label{alg:cascade_train_phase}
\begin{algorithmic}[1]
\State $N_{pool} \gets$ Normal example pool, $P_{train} \gets$ Training set of $N_p$ perturbed examples, $L \gets$ Number Of convolutional layers
\While{current layer $\leq$ $L$ \textbf{Or} $N_{pool} \neq \emptyset$}
\State Draw $N_p$ sized subset $P_{normal}$ from $N_{pool}$
\State $T \gets P_{normal} \cup P_{train}$
\State Train SVM on $T$
\State Predict SVM on $N_{pool}$, eliminate those predicted as normal above a threshold (described in text)
\EndWhile
\end{algorithmic}
\end{algorithm}

The overall false positive rate of a $K$ stage cascade classifier can be represented as:
$F = \prod_{i=1}^K f_i$, where $f_i$ is the false positive rate at each layer. And similarly the true positive rate can be represented in the same form:
$T = \prod_{i=1}^K t_i$ where $t_i$ is the true positive rate at each stage. In order to maximize recall, we maintain a high true positive rate and select a classification threshold which corresponds to a high true positive rate ($97\%$ in AlexNet and $98\%$ in VGG).

\section{Related Work}
\begin{figure}[!h]
\begin{center}
\includegraphics[width=190pt]{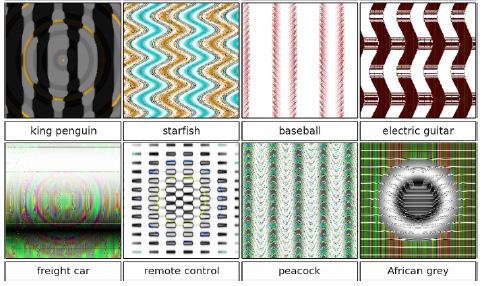}
\vskip -0.3in
\end{center}
\caption{Evolved Adversarial Images Unrecognizable to Humans} \label{fig:fooling_anh}
\vskip -0.1in
\end{figure}

Szegedy et al. \cite{Szegedy2013} proposes the adversarial optimization formulation in eq. (\ref{eqn:adversarial_optimization}). \cite{Goodfellow2014} proposes an explanation of the adversarial mechanism, and proposed a simpler adversarial optimization mechanism that only corrupts based on the signs of gradient of the network. The fact that such examples can be generated so easily with the gradient sign method shows that adversarial examples come from attacking the magnifying effect coming from the linearities in the network. \cite{Nguyen2015} proposes another mechanism to generate adversarials using evolutionary optimization. The result of these do not resemble natural images but still can be classified by deep networks with high confidence(Fig.~\ref{fig:fooling_anh}). \cite{DBLP:journals/corr/Moosavi-Dezfooli15} proposes another efficient approach. \cite{sabour2016adversarial} proposes an approach to generate adversarials that match the convolutional filter outputs as well as perturbing the data. \cite{shaham2015understanding,huang2016learning} propose approaches to sample adversaries or minimax optimization for making learning more robust. While most of the work are done on standard benchmarks such as MNIST, CIFAR and ImageNet, \cite{kurakin2016adversarial} is an interesting work on projecting the adversaries in physical world.

Recently, there have been a lot of focus on training adversarial generation networks to create Generative Adversarial Networks (GANs)~\cite{goodfellow2014generative,radford2015dcgan,zhao2016energy,salimans2016improved}. These networks play a two-player game where a generator network aims to generate adversarials that will not be correctly classified by another discriminator network, and the goal is to generate images more and more similar to natural images. It has been shown that these networks generate images that resemble natural images. However, this generative approach is different from our goal, where we aim to create discriminative networks that discriminates from images that are already indistinguishable from natural images (e.g. Fig.\ref{fig:hard_negatives}).

Mechanisms for detecting and countering adversarial examples have also been proposed~\cite{gu2014towards, papernot2015distillation}. \cite{Luo2016} proposes to use the foveation mechanism to alleviate adversarial examples when it is already known to be adversarial, but did not attempt to detect adversarials. The open-set deep networks proposed by \cite{Bendale2016} seek to alleviate concerns from a soft-max classification by creating an abstain option. The universum classifier~\cite{zhang2015universum} is similar but with more theoretical arguments.

Self-aware learning (classification with an abstain option) had been proposed in e.g.~\cite{kleinberg2010regret,li2011knows}. It is relevant to robust learning (e.g.~\cite{Lanckriet2002}), however robust learning usually seek to directly optimize the minimax loss under adversarial conditions, instead of outputting an abstain option. 
\cite{balsubramani2016learning,wiener2011agnostic} also focuses on classification with an abstain option.

\section{Experiments} \label{sec:experiment}
Our algorithm is tested on 2 approaches to generate adversaries. The main one is data generated using the L-BFGS algorithm by \cite{Szegedy2013}. We generated $4,000$ adversarials from a random subset of the ILSVRC-2012 validation set (total of $50,000$ images). In order to test the out-of-sample generalization capability, we included another dataset, which includes $5,000$ EA-adversarial images generated using the algorithm in~\cite{Nguyen2015}. These datasets are tested with 2 different network structures, VGG-16~\cite{Simonyan2014a} and AlexNet~\cite{Krizhevsky2012}. All input images had been normalized and re-shaped into $224 \times 224$ for VGG-16 model and $227 \times 227$ for AlexNet. The MatConvNet toolbox is used for the deep networks. 
Features are extracted from the statistics of interest, mentioned in section \ref{sec:features}, in the lower $9$ convolutional layers from the VGG-16 network, and in all $5$ convolutional layers from AlexNet.

All the classifiers were trained with a subset of $2,000$ L-BFGS adversarials and a randomly selected subset of $45,000$ normal images in the ILSVRC-2012 dataset. Testing was done on a random held-out dataset with the other $5,000$ normal images and $2,000$ L-BFGS adversarials for the L-BFGS experiments, and on a randomly selected $5,000$ normal images and $5,000$ EA-adversarials for the EA experiments. 
The training and testing sets do not overlap in any means: for all the testing LBFGS adversarials, the corresponding normal images were not seen in the training set.
EA experiments are only conducted with AlexNet since we found out that EA-adversarials did not succeed fooling the VGG-16 network. We have also tried some other fast adversarial generation approaches such as~\cite{Goodfellow2014}, but they generally do not generate adversarials that fool classifiers as completely (with more than $90\%$ confidence) or as universally (on $100\%$ of the images) as the L-BFGS algorithm.

Additional experiment results using DeepFool and on ResNet can be found in the supplementary material.

\subsection{Single-Layer Results}
The first experiments we did were by extracting statistics on each single convolutional layer output, instead of using the cascade classifier. From Tables \ref{tbl:result_Alex} and \ref{tbl:result}, one can see that the statistics from a single layer have some discriminative power to distinguish normal examples from adversarials, but are not extremely effective. However, EA-adversarials were much easier to distinguish, even though our classifier was trained only on L-BFGS adversarials instead of EA ones (Table \ref{tbl:result_Alex_EA}). We only need the first three convolutional layers to reach an overall $97.34\%$ classification accuracy. We believe the reason is that our features capture natural image statistics, and because EA-adversarials look so unnatural, their statistics are vastly different than natural images (see Sec.~\ref{sec:experiment_discussion} for more discussions).

\begin{table}[H]
\begin{center}
\caption{Classification Result with AlexNet for Normal vs. LBFGS-adversarials} \label{tbl:result_Alex}
\begin{tabular}{|c| c c c|}
\hline 
Network Layer & 2nd & 3rd & 4th\\
\hline
Accuracy & $57.5 \pm 0.7$ & $67.3 \pm 0.7$ & $70.9 \pm 0.6$ \\
\hline
Network Layer & 5th & 6th &\\
\hline
Accuracy & $74.9 \pm 0.9$ & $78.95 \pm 0.6$ & \\
\hline
\end{tabular}
\end{center}
\vskip -0.15in
\end{table}

\begin{table}[hbt]
\begin{center}
\caption{Classification Result with VGG-16 for Normal vs. LBFGS-Adersarials} \label{tbl:result}
\begin{tabular}{|c| c c c|}
\hline 
Network Layer & 2nd & 3rd & 4th \\
\hline
Accuracy & $72.1 \pm 0.7$ & $84.1 \pm 0.7$ & $80.3 \pm 0.6$ \\
\hline
Network Layer & 5th & 6th & 7th \\
\hline
Accuracy & $81.4 \pm 0.9$ & $74.3 \pm 0.6$ & $73.9 \pm 0.6$ \\
\hline
Network Layer & 8th & 9th & 10th \\
\hline
Accuracy &  $74.2 \pm 0.7$ & $71.2 \pm 0.7$ & $74.3 \pm 0.8$  \\
\hline
\end{tabular}
\end{center}
\vskip -0.15in
\end{table}

\begin{table}[hbt]
\begin{center}
\caption{Classification Result for Normal vs. EA-Adversarials} \label{tbl:result_Alex_EA}
\begin{tabular}{|c| c c c |}
\hline 
Layer & 2nd & 3rd & 4th\\
\hline
 Accuracy & $93.45 \pm 0.69$ & $98.3 \pm 0.73$ & $97.9 \pm 0.57$\\
\hline
\end{tabular}
\end{center}
\vskip -0.15in
\end{table}
\subsection{Experiment for LBFGS-Adversarials Detection}
\begin{figure*}[htb]
\begin{tabular}{ccc}
\hspace{-0.1in}
\includegraphics[width=0.35\linewidth]{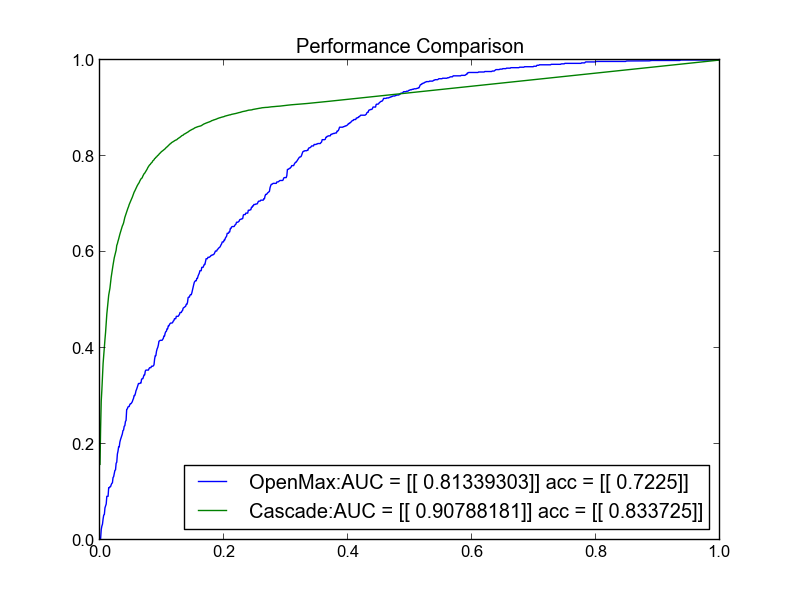}
& 
\hspace{-0.4in}
\includegraphics[width=0.35\linewidth]{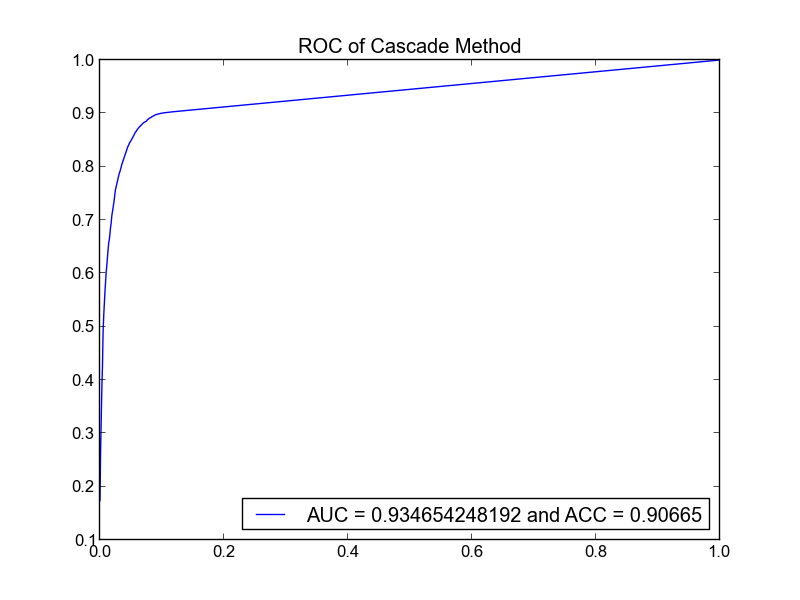}
&
\hspace{-0.4in}
\includegraphics[width=0.35\linewidth]
{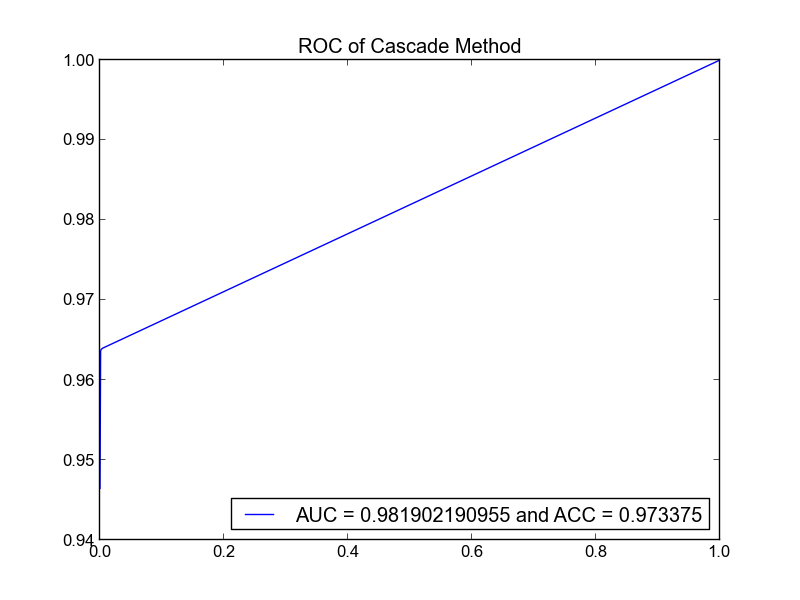}\\
(a) & (b) & (c)
\end{tabular}
\caption{(a) Comparison Between OpenMax detection Methods and Cascade Classifier: The blue curve represents the performace of OpenMax Method, and green curve represents the perfornace for Cascade Classifier.(b) Overall ROC Performance Curve of Cascade Classifier Trained on VGG-16 Network. (c) Overall ROC of data generated from EA-adversarials dataset on AlexNet.} \label{fig:overall_roc}
\vskip -0.1in
\end{figure*}

Next we test the cascade classifier on both AlexNet and VGG-16. The parameter $C$ is set to $0.005$. On AlexNet, the average accuracy of the cascade classifier reaches $83.4\%$ over $20$ random trials, and the AUC (area-under-curve) metric is $90.7\%$. We compared against the recently published OpenMax method~\cite{Bendale2016}. To learn the Weibull distribution required for OpenMax, the EVT was applied on the same training set as the algorithm. Figure \ref{fig:overall_roc}(a) shows the results, where we were able to outperform OpenMax by over $9\%$ in area-under-curve (AUC) and $11\%$ in terms of accuracy.

In VGG-16, the results were even better. The accuracy of the classifier was on average $90.665\%$ over $20$ random trials. Fig.\ref{fig:overall_roc}(b) shows the ROC curve. We believe the fact that VGG has a lot more layers than AlexNet helps setting more constraints on the layer statistics, and is subsequently helpful for detecting adversarial examples. 

Finally, the cascade classifier was tested on EA-adversarials. We obtained more than $96\%$ accuracy with $0$ false positive rate, with a final accuracy of $97.3\%$ and AUC of $98.2\%$ (Fig.\ref{fig:overall_roc}(c)). In other words, our algorithm is rarely fooled by EA-adversarials, even without training on them.

%
%
%

\subsection{Visualization of Statistics} \label{sec:experiment_discussion}

Our experiment results show that EA-adversarials are easy to detect with our detector. To gain more insight into this result, we made a few comparisons between the statistics of interest extracted from normal images, LBFGS-adversarials and EA-adversarials.

\begin{figure*}[htb]
\begin{tabular}{ccc}
\includegraphics[width=0.34\linewidth]{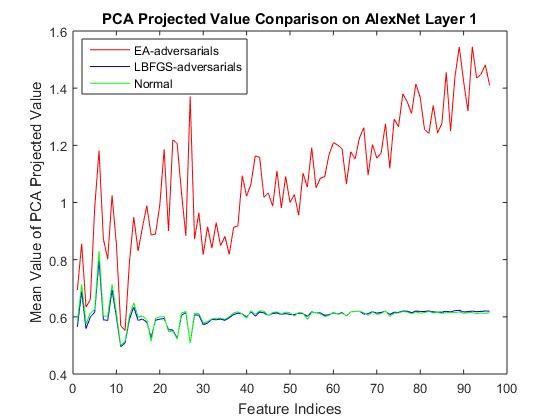}
& 
\hspace{-0.3in}
\includegraphics[width=0.34\linewidth]{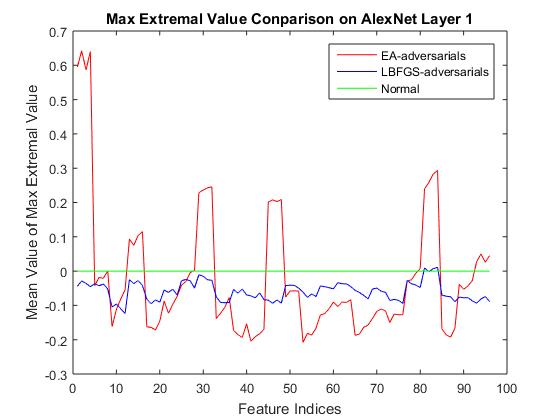}
&
\hspace{-0.3in}
\includegraphics[width=0.34\linewidth]{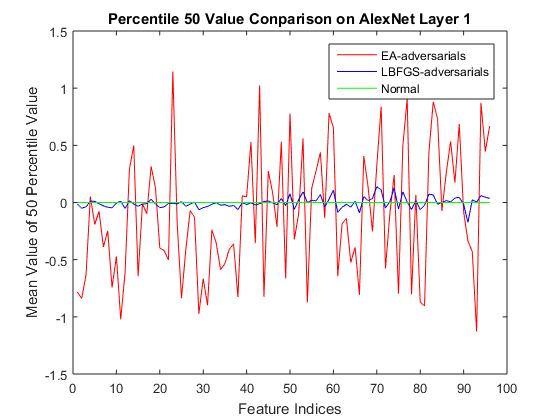} \\
(a) & (b) & (c)
\end{tabular}
\caption{(a) PCA Projection Comparison; (b) Maximum Feature Map Extremal Value Comparison; (c) Median Value Comparison} \label{fig:projected_layer1_alex}
\vskip -0.05in
\end{figure*}

We visualized the average of the statistics that are used for the detection task from the first layer of the AlexNet on all its dimensions. As can be seen in Fig.\ref{fig:projected_layer1_alex}(a), the difference on the PCA projection statistics on  extracted from EA-adversarials and that of the normal images is very dramatic. Meanwhile, compared to the EA-adversarials, the statistics from LBFGS-adversarial have much less difference from the normal data and the difference does not change very much across different dimensions. 

From Fig.~\ref{fig:projected_layer1_alex}(b), one can see that LBFGS-adversarials have smaller extremal values than normal images. This might imply that the LBFGS optimization worked to diminish strong signals from the original image by introducing small pixel perturbations, and that helped our classifiers separating them from normal images. 
From Fig.~\ref{fig:projected_layer1_alex}(c), we see the EA-adversarials evidently differ from normal images. Those results illustrate why EA-adversarials are easier to detect. We suspect it would be easy to reach $100\%$ accuracy, had we actually trained on some EA-adversarials. The capability to generalize to EA-adversarials without training on them showed the general capability of our cascade classifiers to capture natural image statistics and distinguish natural images from unnatural ones.

%
%


\section{Discussions}
\subsection{Self-Aware Learning with an Abstain Option}
The framework of self-aware learning~\cite{li2011knows,Bendale2016,zhang2015universum} considers the case where the learning algorithm has an abstain option of saying ``I don't know'', instead of always making an actual prediction. We define a framework that is slightly different than~\cite{li2011knows}, avoiding the requirement in some frameworks of never making a mistake.

We assume that the training input is drawn i.i.d. from a distribution $P(\mathbf{x},y)$, where $\mathbf{x}$ is the input and $y$ is the output. Assume that the testing input is drawn from a mixture distribution between $P(\mathbf{x},y)$ and $Q(\mathbf{x},y)$:
\begin{equation}
P_m = \Omega P(\mathbf{x},y) + (1-\Omega) Q(\mathbf{x},y)
\end{equation}
, where $\Omega \in \{0,1\}$ is an unknown mixture weight, and $Q(\mathbf{x},y)$ is an adversarial distribution. Assume that we have a classifier that includes a function $f(\mathbf{x})$, and a boolean strategy $a_i$ between \texttt{predict} and \texttt{abstain} that can be chosen for each individual $\mathbf{x}_i$. Assume that the expected error from our classifier on the adversarial distribution is $e_q$ (which could be assumed, if no other prior is present, as the random guessing error of $\frac{C-1}{C}$ for a $C$-class classification problem). Further assume that abstaining always incur a fixed cost $e_a$. As long as $e_a < e_q$, abstaining would be better than predicting on the example drawn from the adversarial distribution, however, $e_a$ should be set sufficiently large so that the classifier would still make predictions when confident, instead of abstaining everything.

For each testing input, the testing of the self-aware classifier is then trying to optimize $\min_a E_{P_m} L_a(\mathbf{x}, y)$ where
\begin{equation}
L_a(\mathbf{x}_i,y_i) = \left\{ \begin{array}{ll}
P(y_i \neq f(\mathbf{x}_i)), & \textrm{if  } a_i = \mathtt{predict},\\
 & (\mathbf{x}_i, y_i) \sim P(\mathbf{x},y) \\ e_q & \textrm{if  } a_i = \mathtt{predict}\\
 &, (\mathbf{x}_i,y_i) \sim Q(\mathbf{x},y)\\e_a & \textrm{if  } a_i = \mathtt{abstain} \end{array} \right.
\end{equation}
hence the classifier needs to select between making a prediction using its function $f(\mathbf{x})$ and risk paying $e_q$ versus abstaining. It is easy to derive the optimal strategy:
\begin{eqnarray}
a_i = \mathtt{predict}, & \textrm{if  } P(\Omega = 1|\mathbf{x_i}) P(y_i \neq f(\mathbf{x_i})) \\
& + P(\Omega = 0|\mathbf{x_i}) e_q < e_a \nonumber \\
a_i = \mathtt{abstain}, & \textrm{otherwise}
\end{eqnarray}
Our approach can be seen as estimating $P(\Omega=1|x_i)$ in this framework. Experiments about the effect of such self-aware learning is shown in the supplementary material. We eagerly hope to apply it in realistic applications in future work.
\subsection{Image Recovery}
Insights from~\cite{Goodfellow2014} indicate that the adversarial mechanism is very specifically attacking vulnerable gradients starting from the first convolutional layer. Insights from the previous experiments also suggest that LBFGS-adversarials work to diminish filter responses from the first convolutional layer. Therefore a natural idea would be to destroy the adversarial effects in the first convolutional layer to try to recover the original image. We tried a very simple approach: applying a small (e.g. $3 \times 3$) average filter on the adversarial image before using the CNN to classify it. The positive and negative adverse gradients will average out in this approach, and make the masked activations from the normal images more prominent. In Table~\ref{tbl:recovery} we illustrate such recovery results: after using a $3\times 3$ average filter on identified adversarial examples, the classification accuracy improved from almost $0\%$ to $73.0\%$, showcasing the effectiveness of this simple average filter.

\begin{table}[htb]
\begin{center}
\caption{Recovery Results. Simply using a $3 \times 3$ average filter we can recover a large proportion of adversarial examples after detecting them using the algorithm described previously. More complex cancellation approaches such as foveation in ~\cite{Luo2016} that utilizes cropping can achieve better results.}
\label{tbl:recovery}
\vskip -0.05in
\begin{tabular}{|c|c|}
\hline
Approach & Top-5 Accuracy \\
& (Recovered Images) \\
\hline
Original Image (Non-corrupted) & $86.5\%$ \\
$3 \times 3$ Average Filter & $73.0\%$ \\
$5 \times 5$ Average Filter & $68.0\%$ \\
Foveation (Object Crop MP)~\cite{Luo2016} & $82.6\%$ \\
\hline
\end{tabular}
\end{center}
\vskip -0.1in
\end{table}

Those results show that we can both detect and recover from adversarial examples with high accuracy. But the main reason we performed this (overly simplistic) experiment is to show how simple it might be to cancel out some adversarial perturbations. Importantly, this result indicates that current deep convolutional networks are too locally focused: these are corruptions that can be cancelled out by a simple $3 \times 3$ average filter, however they can adversely impact the entire result of the deep network. For human with a large receptive field, they will not even care about what happens within a $3 \times 3$ area. Therefore, we believe that future deep learning approaches should focus on enlarging the receptive field in order to reduce the chance of being fooled by adversarial examples. Another potential direction is to research classification approaches that do not require a softmax-type normalization, in order to avoid regularizing attacks such as the ones used in the adversarial optimization in (\ref{eqn:adversarial_optimization}).


\section{Conclusion}
This paper proposes an approach that detects adversarial examples using simple statistics on convolutional layer outputs. A cascade classifier was designed based on simple statistics on filter outputs from each layer. And it was capable of detecting more than $85\%$ of the adversarial examples. Experiments showed that our cascade classifier significantly outperforms state-of-the-art on detecting adversarial examples. Experiment also showed transfer learning capabilities of our classifier, since the classifier we trained with L-BFGS adversarials are capable of detecting EA-adversarials as well. Insights drawn from these experiments lead us to perform simple $3\times 3$ average filter to corrupted images, which successfully recovered most of them. In the future, we would like to explore GAN-type generative adversarial networks from the current results, with multiple rounds of adversarial detection and counter-detection.

\subsubsection*{Acknowledgements}
This paper was supported by Future of Life grants 2015-143880 and 2016-158701.

{\small
\bibliographystyle{ieee}
\bibliography{dependent_reference}

\begin{thebibliography}{10}\itemsep=-1pt

\bibitem{balsubramani2016learning}
A.~Balsubramani.
\newblock Learning to abstain from binary prediction.
\newblock {\em arXiv preprint arXiv:1602.08151}, 2016.

\bibitem{Bendale2016}
A.~Bendale and T.~E. Boult.
\newblock Towards open set deep networks.
\newblock In {\em IEEE Conference on Computer Vision and Pattern Recognition},
  2016.

\bibitem{goodfellow2014generative}
I.~Goodfellow, J.~Pouget-Abadie, M.~Mirza, B.~Xu, D.~Warde-Farley, S.~Ozair,
  A.~Courville, and Y.~Bengio.
\newblock Generative adversarial nets.
\newblock In {\em Advances in Neural Information Processing Systems}, pages
  2672--2680, 2014.

\bibitem{Goodfellow2014}
I.~J. Goodfellow, J.~Shlens, and C.~Szegedy.
\newblock Explaining and harnessing adversarial examples.
\newblock {\em arXiv preprint arXiv:1412.6572}, 2014.

\bibitem{gu2014towards}
S.~Gu and L.~Rigazio.
\newblock Towards deep neural network architectures robust to adversarial
  examples.
\newblock {\em arXiv preprint arXiv:1412.5068}, 2014.

\bibitem{Hastie2001}
T.~Hastie, R.~Tibshirani, and J.~Friedman.
\newblock {\em The Elements of Statistical Learning}.
\newblock Springer-Verlag, New York, 2001.

\bibitem{He2016residual}
K.~He, X.~Zhang, S.~Ren, and J.~Sun.
\newblock Deep residual learning for image recognition.
\newblock In {\em IEEE Conference on Computer Vision and Pattern Recognition},
  2016.

\bibitem{huang2016learning}
R.~Huang, B.~Xu, D.~Schuurmans, and C.~Szepesv{\'a}ri.
\newblock Learning with a strong adversary.
\newblock In {\em International Conference on Learning Representations}, 2016.

\bibitem{Indyk1998}
P.~Indyk and R.~Motwani.
\newblock Approximate nearest neighbors: towards removing the curse of
  dimensionality.
\newblock In {\em Proceedings of the thirtieth annual ACM symposium on Theory
  of computing}, pages 604--613, 1998.

\bibitem{ioffe2015batch}
S.~Ioffe and C.~Szegedy.
\newblock Batch normalization: Accelerating deep network training by reducing
  internal covariate shift.
\newblock {\em arXiv preprint arXiv:1502.03167}, 2015.

\bibitem{PCA}
I.~Jolliffe.
\newblock {\em Principle Component Analysis}.
\newblock Springer-Verlag, 1986.

\bibitem{kleinberg2010regret}
R.~Kleinberg, A.~Niculescu-Mizil, and Y.~Sharma.
\newblock Regret bounds for sleeping experts and bandits.
\newblock {\em Machine learning}, 80(2-3):245--272, 2010.

\bibitem{Krizhevsky2012}
A.~Krizhevsky, I.~Sutskever, and G.~E. Hinton.
\newblock Imagenet classification with deep convolutional neural networks.
\newblock In {\em Advances in Neural Information Processing Systems}, pages
  1097--1105, 2012.

\bibitem{kurakin2016adversarial}
A.~Kurakin, I.~Goodfellow, and S.~Bengio.
\newblock Adversarial examples in the physical world.
\newblock {\em arXiv preprint arXiv:1607.02533}, 2016.

\bibitem{Lanckriet2002}
G.~R. Lanckriet, L.~E. Ghaoui, C.~Bhattacharyya, and M.~I. Jordan.
\newblock A robust minimax approach to classification.
\newblock {\em Journal of Machine Learning Research}, 3:555--582, 2003.

\bibitem{li2011knows}
L.~Li, M.~L. Littman, T.~J. Walsh, and A.~L. Strehl.
\newblock Knows what it knows: a framework for self-aware learning.
\newblock {\em Machine learning}, 82(3):399--443, 2011.

\bibitem{li2017filter}
X.~Li, F.~Li, X.~Fern, and R.~Raich.
\newblock Filter shaping for convolutional networks.
\newblock In {\em International Conference on Learning Representations}, 2017.

\bibitem{Luo2016}
Y.~Luo, X.~Boix, G.~Roig, T.~A. Poggio, and Q.~Zhao.
\newblock Foveation-based mechanisms alleviate adversarial examples.
\newblock {\em arXiv preprint arXiv:1511.06292v3}, 2016.

\bibitem{DBLP:journals/corr/Moosavi-Dezfooli15}
S.~Moosavi{-}Dezfooli, A.~Fawzi, and P.~Frossard.
\newblock Deepfool: a simple and accurate method to fool deep neural networks.
\newblock {\em CoRR}, abs/1511.04599, 2015.

\bibitem{Nguyen2015}
A.~Nguyen, J.~Yosinski, and J.~Clune.
\newblock Deep neural networks are easily fooled: High confidence predictions
  for unrecognizable images.
\newblock In {\em IEEE Conference on Computer Vision and Pattern Recognition},
  2015.

\bibitem{papernot2015distillation}
N.~Papernot, P.~McDaniel, X.~Wu, S.~Jha, and A.~Swami.
\newblock Distillation as a defense to adversarial perturbations against deep
  neural networks.
\newblock {\em arXiv preprint arXiv:1511.04508}, 2015.

\bibitem{radford2015dcgan}
A.~Radford, L.~Metz, and S.~Chintala.
\newblock Unsupervised representation learning with deep convolutional
  generative adversarial networks.
\newblock {\em arXiv preprint arXiv:1511.06434}, 2015.

\bibitem{sabour2016adversarial}
S.~Sabour, Y.~Cao, F.~Faghri, and D.~J. Fleet.
\newblock Adversarial manipulation of deep representations.
\newblock In {\em International Conference on Learning Representations}, 2016.

\bibitem{salimans2016improved}
T.~Salimans, I.~Goodfellow, W.~Zaremba, V.~Cheung, A.~Radford, and X.~Chen.
\newblock Improved techniques for training gans.
\newblock {\em arXiv preprint arXiv:1606.03498}, 2016.

\bibitem{shaham2015understanding}
U.~Shaham, Y.~Yamada, and S.~Negahban.
\newblock Understanding adversarial training: Increasing local stability of
  neural nets through robust optimization.
\newblock {\em arXiv preprint arXiv:1511.05432}, 2015.

\bibitem{Simonyan2014a}
K.~Simonyan and A.~Zisserman.
\newblock Very deep convolutional networks for large-scale image recognition.
\newblock {\em arXiv preprint arXiv:1409.1556}, 2014.

\bibitem{GoogleLeNet}
C.~Szegedy, W.~Liu, Y.~Jia, P.~Sermanet, S.~Reed, D.~Anguelov, D.~Erhan,
  V.~Vanhoucke, and A.~Rabinovich.
\newblock Going deeper with convolutions.
\newblock {\em arXiv:1409.4842}, 2014.

\bibitem{Szegedy2013}
C.~Szegedy, W.~Zaremba, I.~Sutskever, J.~Bruna, D.~Erhan, I.~Goodfellow, and
  R.~Fergus.
\newblock Intriguing properties of neural networks.
\newblock {\em arXiv preprint arXiv:1312.6199}, 2013.

\bibitem{viola2004robust}
P.~Viola and M.~J. Jones.
\newblock Robust real-time face detection.
\newblock {\em International journal of computer vision}, 57(2):137--154, 2004.

\bibitem{wiener2011agnostic}
Y.~Wiener and R.~El-Yaniv.
\newblock Agnostic selective classification.
\newblock In {\em Advances in Neural Information Processing Systems}, pages
  1665--1673, 2011.

\bibitem{zhang2015universum}
X.~Zhang and Y.~LeCun.
\newblock Universum prescription: Regularization using unlabeled data.
\newblock In {\em AAAI Conference on Artificial Intelligence}, 2017.

\bibitem{zhao2016energy}
J.~Zhao, M.~Mathieu, and Y.~LeCun.
\newblock Energy-based generative adversarial network.
\newblock {\em arXiv preprint arXiv:1609.03126}, 2016.

\end{thebibliography}
}
\newpage
\setcounter{figure}{0}
\setcounter{section}{0}
{\bf \large Supplementary Material}

\section{Results on DeepFool}
For this experiment we used 5000 adversarial images generated with the DeepFool
algorithm. We collected a training set of images using the adversarials, with an equal
number of real images drawn from the ILSVRC2012 validation set. For the Deep-
Fool adversarials we used the implementation given in the Foolbox algorithm library.
Given this data, we chose to use the ResNet 50 architecture as the backbone CNN
for our experiments. Because we use ImageNet data, we preprocessed our training
set accordingly: all images were reshaped to 224x224x3, the channels were modified
to BGR ordering, and the channel-wise mean was subtracted from each sample. We
then performed a forward pass as usual on all images, performing the classifier cascade
at each layer. Given the output of each convolutional layer, we extracted features
that would characterize images as being either from a real or adversarial distribution.
For each output we extracted PCA coefficients, extremal values, and values within
the 25th, 50th and 75th percentile to form a new training sample. We then used an
SVM to learn the real statistics from the adversarials. The real examples were given
a ground truth label of 1, and the adversarials were given a label of 0. We performed
a parameter search over kernel type and C value. In all our experiments a C of 0.005
and a linear kernel performed the best. We then tested on 2000 real images and 2000
DeepFool adversarials. In our experiments with ResNet, we only performed the
cascade for the first three layers before we found nearly all of the adversarial images.

\begin{figure*}[htb]
\includegraphics[width=330pt]{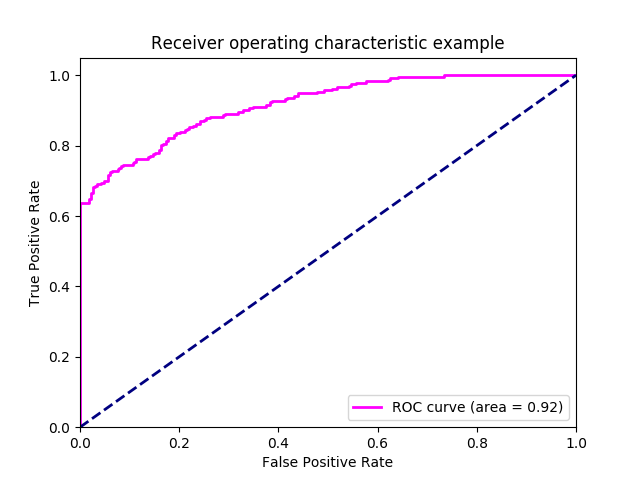}
\caption{ROC Curve for Detecting DeepFool adversarials in  ResNet-50, the algorithm maintained more than $90\%$ in AUC}
\label{fig:deepfool}
\end{figure*}

The result can be seen in Fig.~\ref{fig:deepfool}. The algorithm maintained more than $90\%$ AUC, showing that DeepFool did not fundamentally change the type of adversarials.

\section{Results on Self-Aware Learning}
We implemented self aware learning using ResNet 50, the goal being to use the
softmax probabilities to learn parameters that would cause the network to more intelligently
classify inputs. The network should classify an image if it was sure the
image was real, or abstain if either the network was not sufficiently confident, or if
the image was adversarial, as described in Sec. 7.1. To test the presented algorithm, we use $2,000$ real images
drawn from ILSVRC2012 validation set, and $2,000$ adversarial images from the testing set of the previous experiment, generated using
the DeepFool algorithm. We tested the self aware learning algorithm with a high $e_q = 10$. This worked well enough that the network chose to abstain
or classify, rather than incur a high penalty for guessing incorrectly. 
 We observed that for each testing image, our estimation of the source
distribution resulted in $e_a$ between 2 and 8. We then varied $e_a$ between these values to see if there was a threshold at which we could abstain
from all adversarials, retaining predictions for only real examples. We were also
interested in thresholds that maximized the true positive rate (prediction of real examples)
while abstaining from as many adversarials as possible. We found the lower
thresholds resulted in the abstaining from predicting on all adversarials, but it also
abstained from many (but not all) real examples. Higher thresholds resulted in many
more real predictions retained, but some also some adversarials made it through. High
thresholds would finally result in the network not abstaining at all.

\begin{figure*}[htb]
\includegraphics[width=330pt]{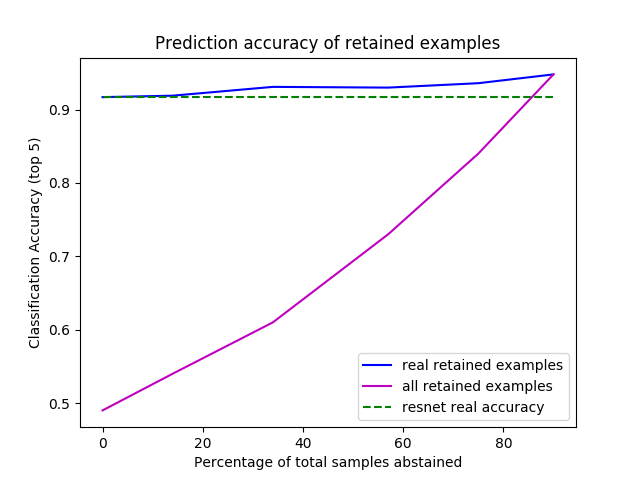}\caption{Self-aware learning results. In a mixture of half real and half adversarial examples, the classification accuracy of discarding nothing falls a little under $50\%$, with more examples abstained, the accuracy improves significantly. The accuracy of retained normal examples (blue curve) also improves when more examples are abstained, as the abstained examples also include normal examples that are not predicted confidently.} \label{fig:self_aware}
\end{figure*}

The results can be seen in Fig.~\ref{fig:self_aware}. It can be seen that besides abstaining adversarial examples, the system also abstains from predicting on some normal examples that the classifier is not confident on. Hence, with a high abstain ratio the prediction accuracy on normal examples is also higher.

\section{Images Classified Correctly and Incorrectly}
In this section we show some images classified correctly and incorrectly from the algorithm. Unfortunately we are not quite able to observe any particular visible trends, maybe due to the subtlety of adversarial images.
\begin{figure*}[h]
\begin{tabular}{cccc}
\includegraphics[width=90pt]{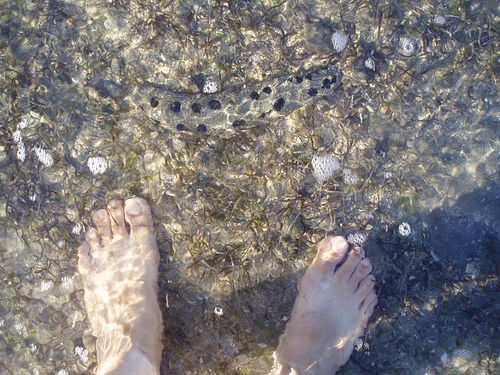} & \includegraphics[width=90pt]{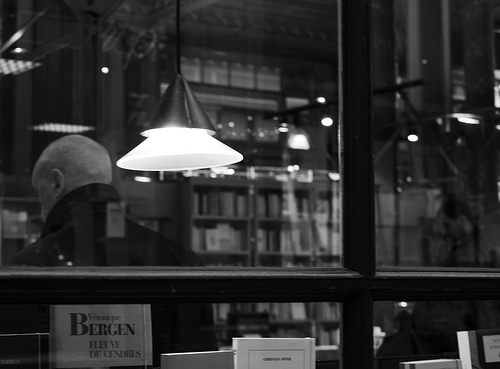} &
\includegraphics[width=90pt]{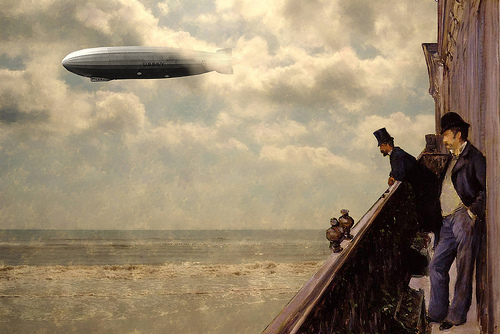} &
\includegraphics[width=90pt]{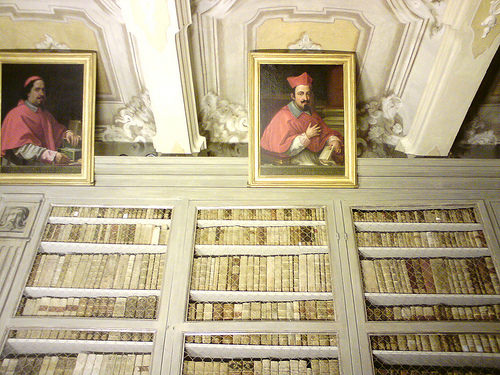} \\
(a) & (b) & (c) & (d)
\end{tabular}
\caption{Some of Misclassification on L-BFGS images by Our Classifier. (a) and (b) are from normal dataset. (c) and (d) are from LBFGS-Adversarial dataset, which is misclassified to category n02408429(water buffalo) and n01518878(ostrich, Struthio camelus).}
\label{fig:classifier_visualization}
\end{figure*}

\begin{figure*}[h]
\begin{tabular}{cccc}
\includegraphics[width=90pt]{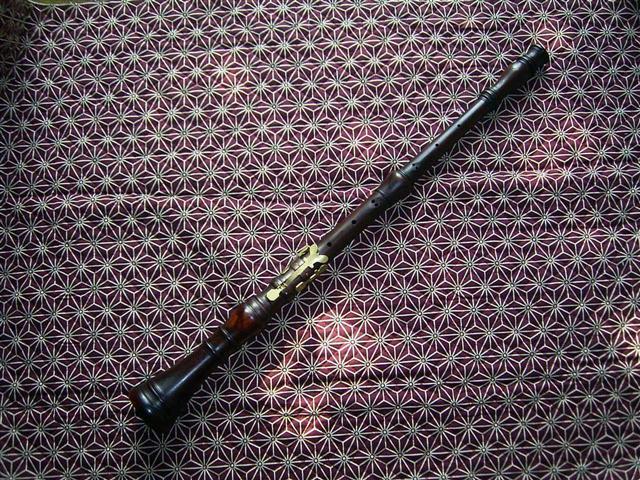} & \includegraphics[width=90pt]{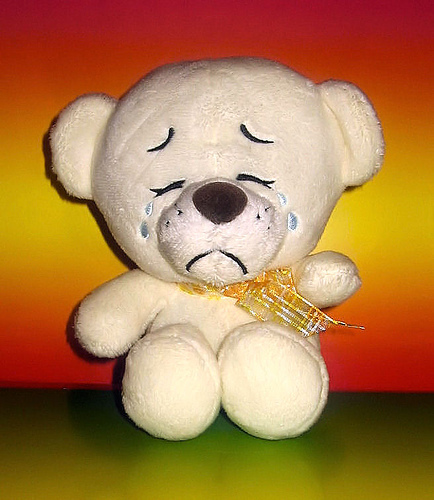} &
\includegraphics[width=90pt]{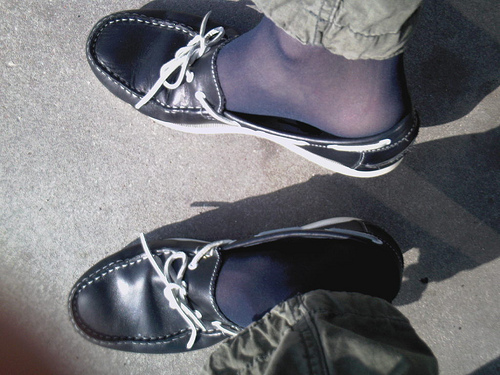} &
\includegraphics[width=90pt]{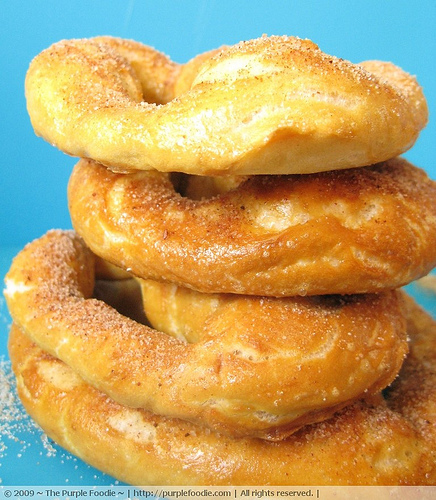} \\
(a) & (b) & (c) & (d)
\end{tabular}
\caption{Some of Correctly Classified on L-BFGS images by Our Classifier. (a) and (b) are from normal dataset. (c) and (d) are from LBFGS-Adversarial dataset, which is misclassified to category n04209133(shower cap) and n02328150(Angora).}
\label{fig:classifier_visualization}
\end{figure*}

\begin{figure*}[h]
\begin{tabular}{cccc}
\includegraphics[width=90pt]{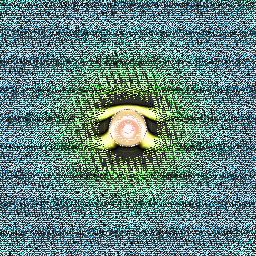} & \includegraphics[width=90pt]{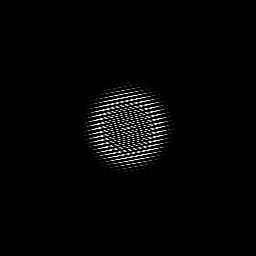} &
\includegraphics[width=90pt]{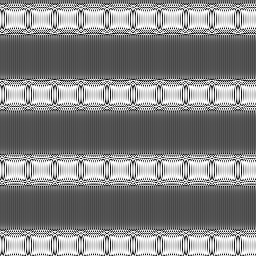} &
\includegraphics[width=90pt]{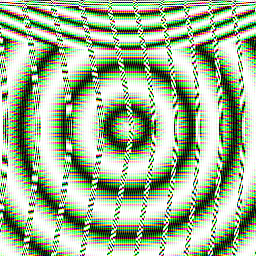} \\
(a) & (b) & (c) & (d)
\end{tabular}
\caption{Some of Misclassfied EA images by Our Classifier. From left to right, they are misclssified to category n03220513 (dome), n01749939 (green mamba), n04118776 (rule, ruler) and n03935335 piggy (bank, penny bank)}
\label{fig:classifier_visualization}
\end{figure*}

\begin{figure*}[h]
\begin{tabular}{cccc}
\includegraphics[width=90pt]{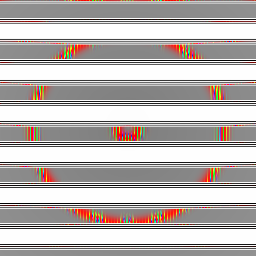} & \includegraphics[width=90pt]{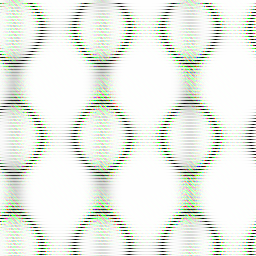} &
\includegraphics[width=90pt]{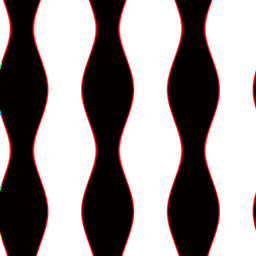} &
\includegraphics[width=90pt]{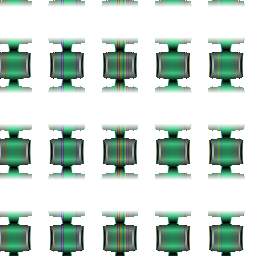} \\
\end{tabular}
\caption{Some of Correctly Classified EA images by Our Classifier. From left to right they are misclassified to n06874185 (traffic light, traffic signal, stoplight), n03443371 (goblet), n04522168 (vase) and n03742115 (medicine chest, medicine cabinet)}
\label{fig:classifier_visualization}
\end{figure*}

\begin{figure*}[h]
\begin{tabular}{cccc}
\includegraphics[width=90pt]{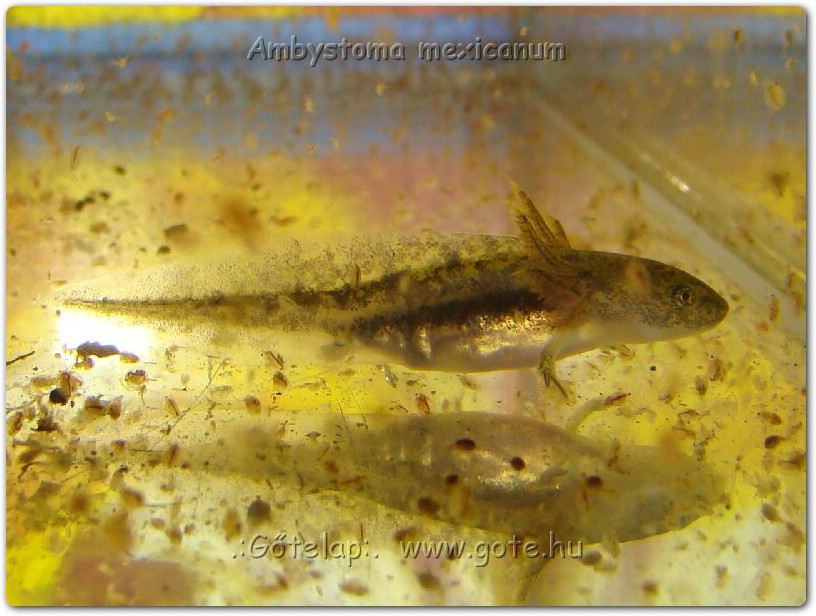} & \includegraphics[width=90pt]{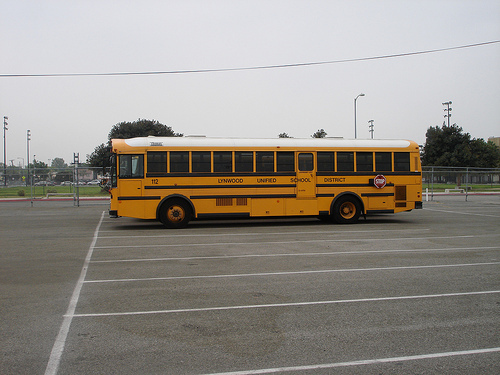} &
\includegraphics[width=90pt]{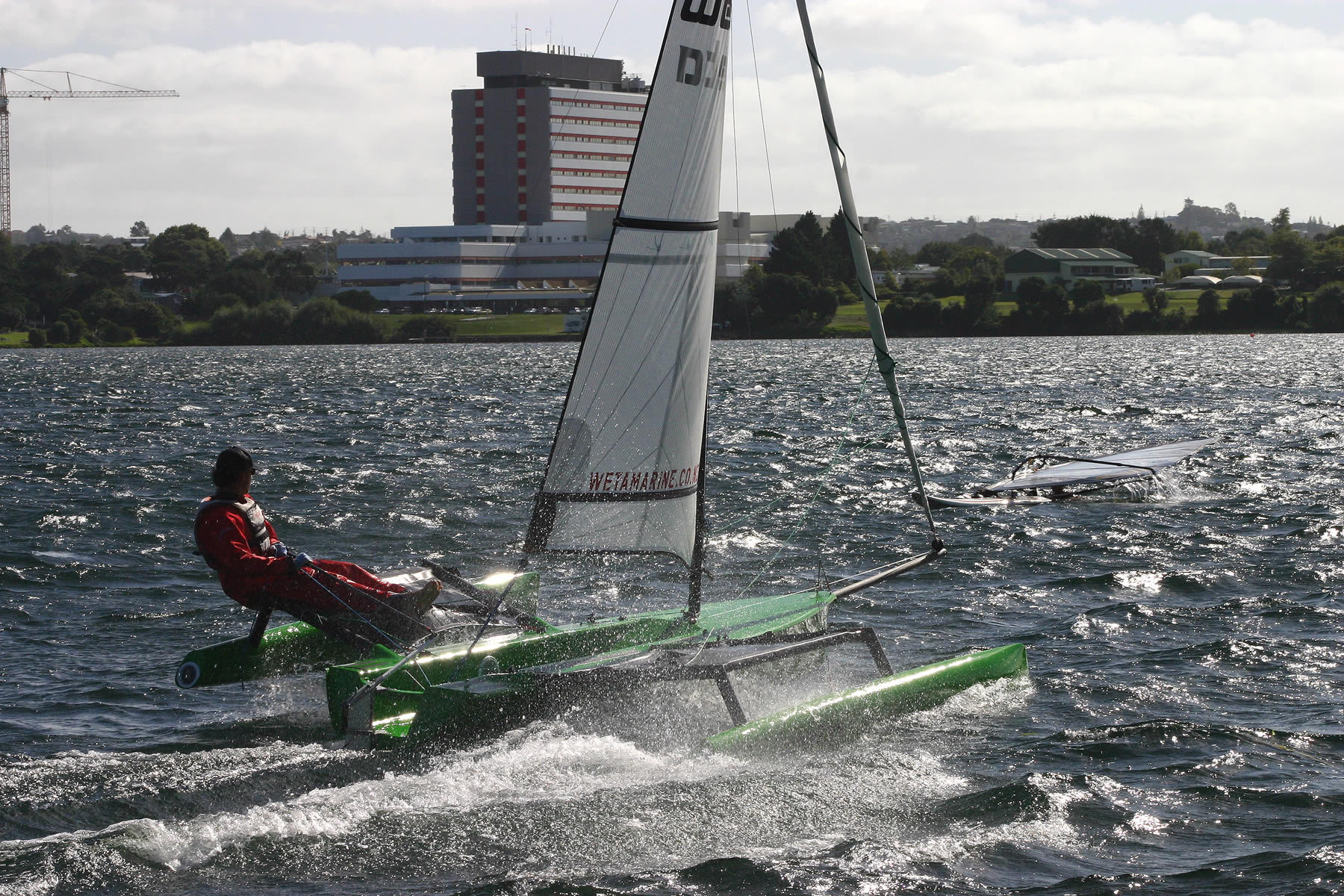} &
\includegraphics[width=90pt]{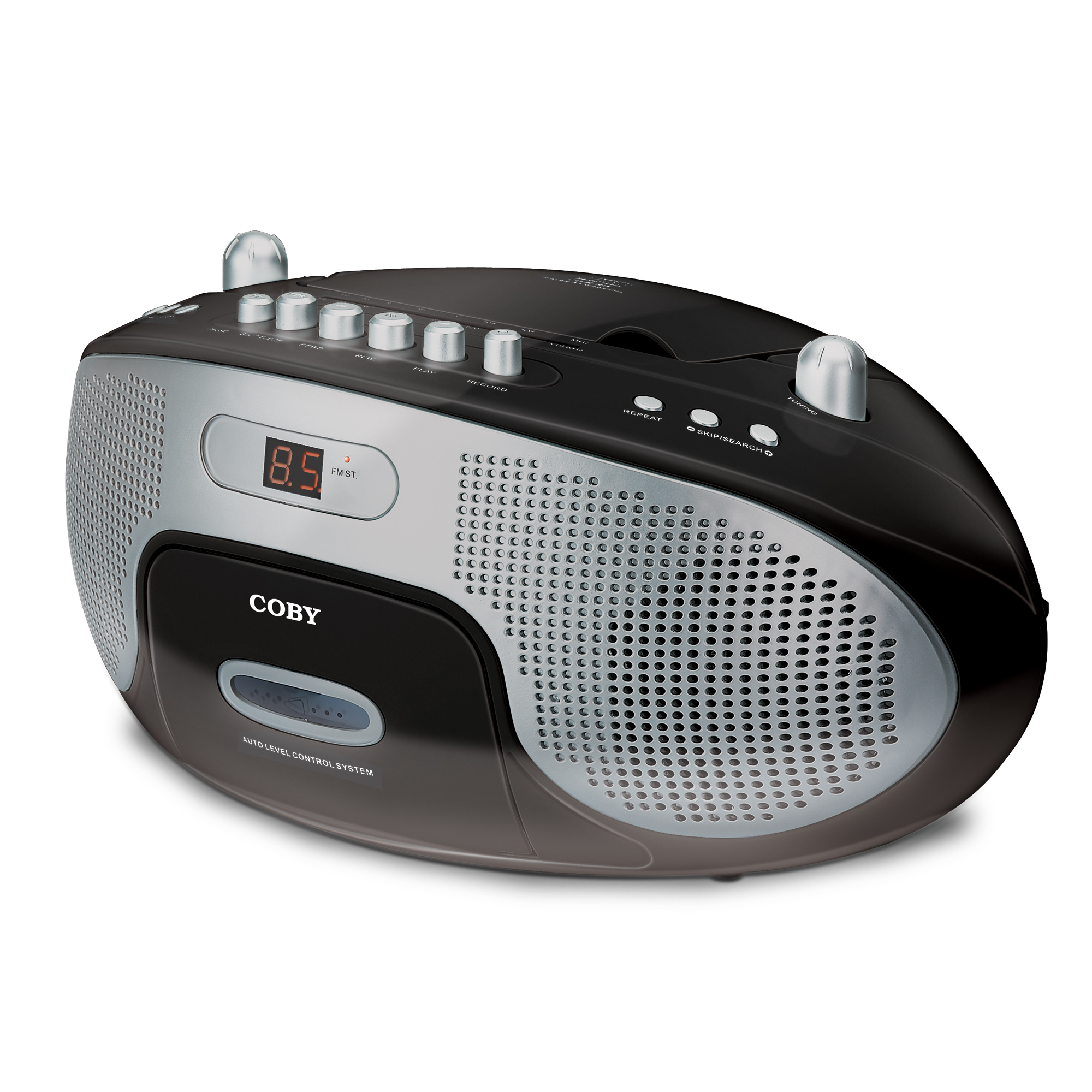} \\
(a) & (b) & (c) & (d)
\end{tabular}
\caption{Images Misclassified by OpenSet Method but Correctly Classified by Our Classifier. (c) and (d) are from LBFGS-Adversarial dataset, which is misclassified to category n02133161(American black bear) and n02328150(Agona).}
\label{fig:classifier_visualization}
\end{figure*}

\end{document}